\documentclass{article}
\usepackage{ijcai19}

\usepackage{times}
\usepackage{soul}
\usepackage{url}
\usepackage[hidelinks]{hyperref}
\usepackage[utf8]{inputenc}
\usepackage[small]{caption}
\usepackage{graphicx}
\usepackage{amsmath}
\usepackage{booktabs}
\usepackage{algorithm}
\usepackage{algorithmic}
\usepackage{times}
\usepackage{latexsym}
\usepackage{url}
\usepackage{makecell}
\usepackage{multirow}
\usepackage{balance}
\usepackage{graphicx}
\usepackage{capt-of}
\usepackage{varwidth}
\usepackage{enumitem}
\newsavebox\tmpbox
\usepackage{subfig}
\usepackage[symbol]{footmisc}

\newcommand{\datasetname}{AmazonQA}
\title{AmazonQA: A Review-Based 
Question Answering Task}

\author{
Mansi Gupta\footnote{contributed equally to this work}\footnote{currently at Petuum, Inc.}\and
Nitish Kulkarni$^{*\ddagger}$\and
Raghuveer Chanda$^*$\footnote{currently at Google LLC}\and \\
Anirudha Rayasam\And
Zachary C Lipton
\affiliations
Carnegie Mellon University
\emails
\{mgupta1410, ni4lgi, raghuveer.chanda, rcanirudha\}@gmail.com, zlipton@cs.cmu.edu
}

\begin{document}

\maketitle

\begin{abstract}

Every day, thousands of customers 
post questions on Amazon product pages.
After some time, if they are fortunate, 
a knowledgeable customer might answer their question.
Observing that many questions can be answered 
based upon the available product reviews, 
we propose the task of review-based QA.
Given a corpus of reviews and a question,
the QA system synthesizes an answer. 
To this end, we introduce a new dataset and propose a method 
that combines information retrieval techniques 
for selecting relevant reviews (given a question) 
and ``reading comprehension'' models 
for synthesizing an answer (given a question and review). 
Our dataset consists of $923$k questions, 
$3.6$M answers and $14$M reviews across $156$k products. 
Building on the well-known Amazon dataset,
we collect additional annotations, marking each question
as either answerable or unanswerable based on the available reviews.
A deployed system could first classify a question as answerable
and then attempt to generate an answer.
Notably, unlike many popular QA datasets,
here the questions, passages, and answers
are all extracted from real human interactions.
We evaluate numerous models for answer generation
and propose strong baselines, 
demonstrating the challenging nature of this new task.

\end{abstract}

\section{Introduction}



E-commerce customers at websites like Amazon
post thousands of product-specific questions per day.
\begin{figure}[ht]
    \centering
    \includegraphics[scale=0.4]{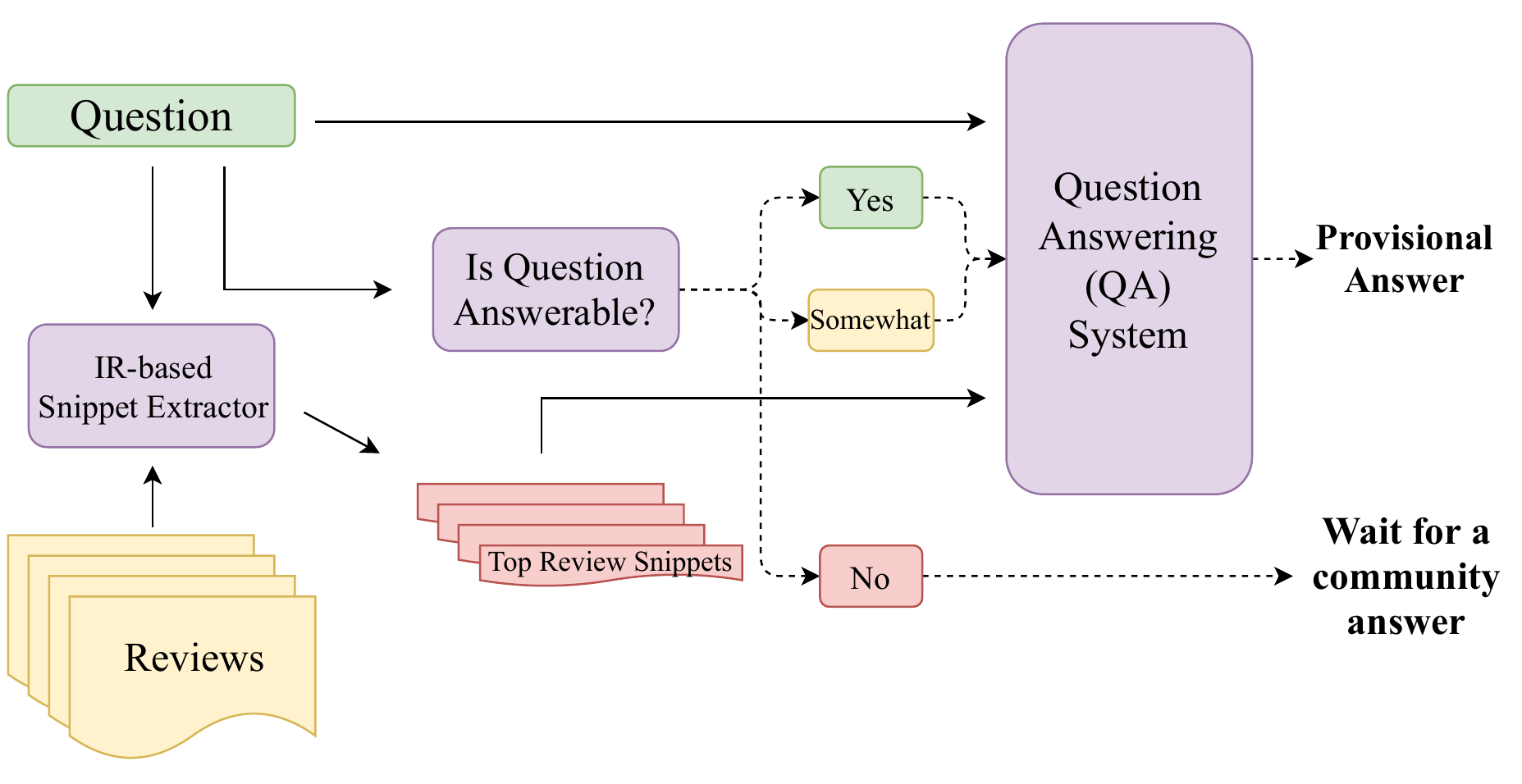}
    \caption{A community question-answering platform that provides provisional answers to questions using reviews.}
    \label{fig:cqa_system}
\end{figure}
From our analysis of a large-scale dataset 
crawled from Amazon's website \cite{mcauley2016addressing},
comprising question-answering data and product reviews, 
we observe that:
(i) most questions have a response time of several days,
with an average of around $2$ days per question
(excluding those questions which remain unanswered indefinitely); 
(ii) the product reviews are comparatively elaborate and informative
as judged against the answers posted to specific questions;
and (iii) following human-in-the-loop experiments,
we discovered that more than half of the questions 
can (at least partially) be answered using existing reviews (\S\ref{sec:answer}). 

Figure \ref{fig:cqa_system} depicts our conception
of a system that can provide on-demand, 
immediate answers to the user questions 
in a community question answering platform, 
by leveraging existing user reviews. 
Motivated by the possibility of designing such systems,
we introduce the \emph{review-based community question answering} task:
\begin{quote}
Given a set of product reviews and a question concerning a specific product,
generate an informative natural language answer.
\end{quote}

We build upon the \emph{question-answering} (QA) 
and product review dataset due to \cite{mcauley2016addressing},
incorporating additional curation and annotations
to create a new resource for automatic community question answering.
Our resource, denoted \emph{AmazonQA}, 
offers the following distinctive qualities: 
(i) it is extracted entirely from existing, real-world data;
and (ii) it may be the largest public QA dataset with descriptive answers. 
To facilitate the training of complex ML-based QA models on the dataset, 
we provide rich pre-processing, extracting top review snippets 
for each question based on information retrieval (IR) techniques, 
filtering outliers and building an \textit{answerability} classifier 
to allow for training of QA models on only the answerable questions.
Further, we implement a number of heuristic-based and neural QA models 
for benchmarking the performance of some of 
the best-performing existing QA models on this dataset. 
We also provide human evaluation by experts 
as well as the users from the community question answering platform. 



\section{Related Work}
\begin{table*}[t!]
    \small{}
    \centering
    \begin{tabular}{l l l l l l} 
        \toprule
        Dataset & Question Source & Document & Answer Type & \# Qs & \# Documents \\ 
         & & Source &  & &  \\ 
        \midrule
        NewsQA \cite{trischler2016newsqa} & Crowd-Sourced & CNN & Span of words & 100K & 10K \\
        SearchQA \cite{dunn2017searchqa} & Generated & WebDoc & Span of words & 140K & 6.9M passages\\
        SQuAD \cite{rajpurkar2016squad} & Crowd-Sourced & Wiki & Span of words & 100K & 536\\
        RACE \cite{lai2017race} & Crowd-Sourced & English Exams & Multiple choice & 97l & 28K\\
        ARC \cite{clark2018think} & Generated & WebDoc & Multiple choice & 7787 & 14M sentences \\
        DuReader \cite{he2017dureader} & Crowd-Sourced & WebDoc/CQA & Manual summary & 200K & 1M \\
        Natural Questions \cite{kwiatkowski2019natural} & User Logs & Wiki & Span of words, entities & 307K & 307K \\
        NarrativeQA \cite{kovcisky2018narrativeqa} & Crowd-Sourced & Books \& Movies & Manual summary & 46.7K & 1,572 stories\\
        MS MARCO \cite{nguyen2016ms} & User Logs & WebDoc & Manual summary & 1M & 8.8M passages, \\
                 &           &   &                & & 3.2M docs\\
        \bottomrule
    \end{tabular}
    \caption{Existing QA Datasets}
    \label{tab:existing_datasets}
\end{table*}

\paragraph{Open-World and Closed-World Question Answering} 
An open-world question-answering dataset constitutes 
a set of question-answer pairs accompanied by a knowledge database, 
with no explicit link between the question-answer pairs and knowledge database entries. SimpleQA~\cite{bordes2015simpleqa} is a representative example of such a dataset. 
It requires simple reasoning over the Freebase knowledge database to answer questions.
In closed-world question answering, 
the associated snippets are sufficient 
to answer all corresponding questions. 
Despite using open-world snippets in \datasetname, 
we pose the final question-answering in a closed-world setting. 
We ensure that for any answerable question,
the associated snippets contain all the required supporting information. 
Below, we highlight key distinguishing features 
of the recent popular closed-world QA datasets,
comparing and contrasting them with \datasetname. 
Basic statistics for related dataset are shown in Table \ref{tab:existing_datasets}.


\paragraph{Span-based Answers} 
SQuAD~\cite{rajpurkar2016squad,rajpurkar2018know} is a single-document dataset. 
The answers are multi-word spans from the context. 
To address the challenge of developing QA systems 
that can handle longer contexts, 
SearchQA \cite{dunn2017searchqa} presents contexts 
consisting of more than one document. 
Here, the questions are not guaranteed 
to require reasoning across multiple documents 
as the supporting documents are collected 
through information retrieval 
after the (question, answer) pairs are determined. 
We follow a similar information retrieval scheme 
to curate \datasetname{}, but unlike \cite{dunn2017searchqa} 
whose answers are spans in the passage,
ours are free-form.
\paragraph{Free-form Answers} 
Some recent datasets, including \cite{nguyen2016ms,he2017dureader} 
have free-form answer generation. 
MS MARCO \cite{nguyen2016ms} contains user queries 
from Bing Search with human generated answers. 
Systems generate free-form answers 
and are evaluated by automatic metrics such as ROUGE-L and BLEU-1. 
Another variant with human generated answers is DuReader \cite{he2017dureader} 
for which the questions and documents 
are based on user queries from Baidu Search and Baidu Zhidao.
\paragraph{Community/Opinion Question Answering} 
The idea of question-answering using reviews and product information 
has been previously explored.
\cite{mcauley2016addressing} address subjective queries 
using the relevance of reviews. 
\cite{wan2016modeling} extend this work 
by incorporating aspects of personalization and ambiguity. 
\cite{yu2012answering} employ SVM classifiers 
for identifying the question aspects, question types, 
and classifying responses as opinions or not,
optimizing salience, coherence and diversity to generate an answer. 
However, to the best of our knowledge, 
no prior work answers user queries
from the corresponding reviews. 
Our baseline models are inspired by machine comprehension models 
like Bi-Directional Attention Flow (BiDAF) 
which predict the start and end positions of spans in the context.




\section{Dataset}
We build upon the dataset of \cite{mcauley2016addressing}, 
who collected reviews, questions, and answers 
by scraping the product pages of Amazon.com, 
for the period of May 1996 to July 2014, 
spanning 17 categories of products including
\emph{Electronics}, \emph{Video Games}, \emph{Home and Kitchen}, etc.
First, we preprocess and 
expand (via new annotations) this raw dataset to suit the QA task,
detailing each step, and characterizing dataset statistics
in the following sections.


\begin{figure*}[ht]
    \centering
    \subfloat{\includegraphics[width=5.2cm]{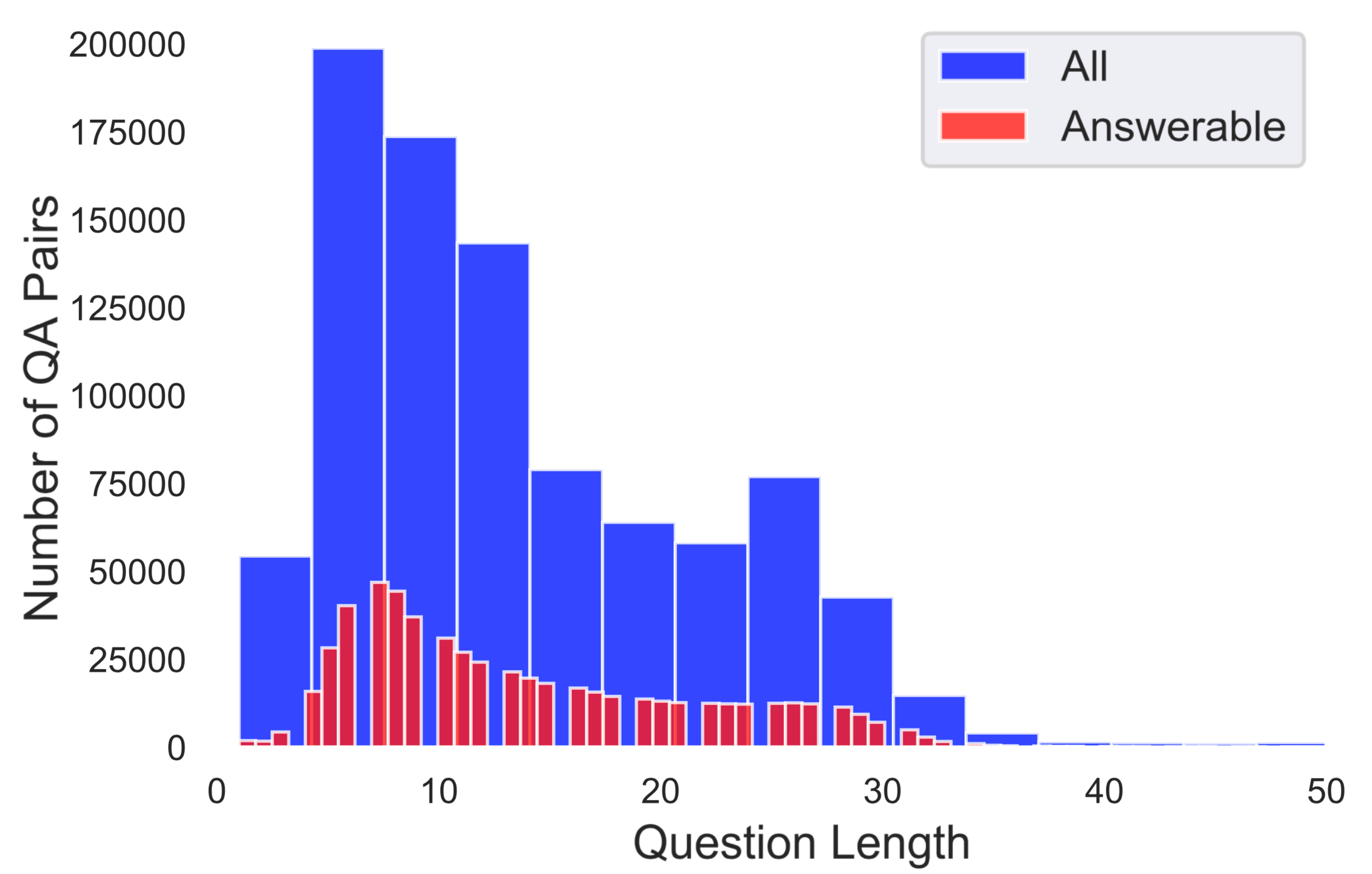} }
    \subfloat{\includegraphics[width=5.2cm]{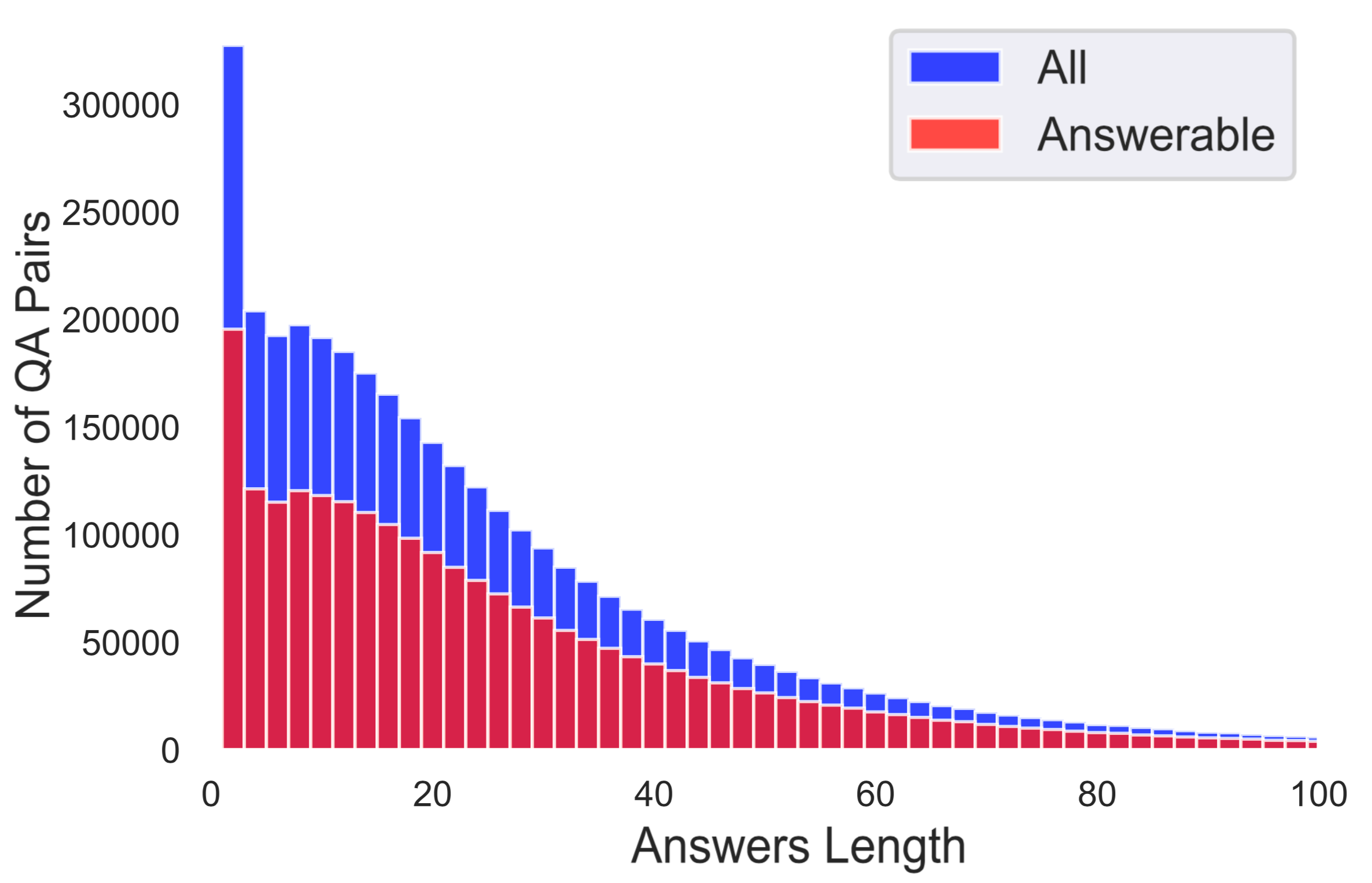} }
    \subfloat{\includegraphics[width=5.2cm]{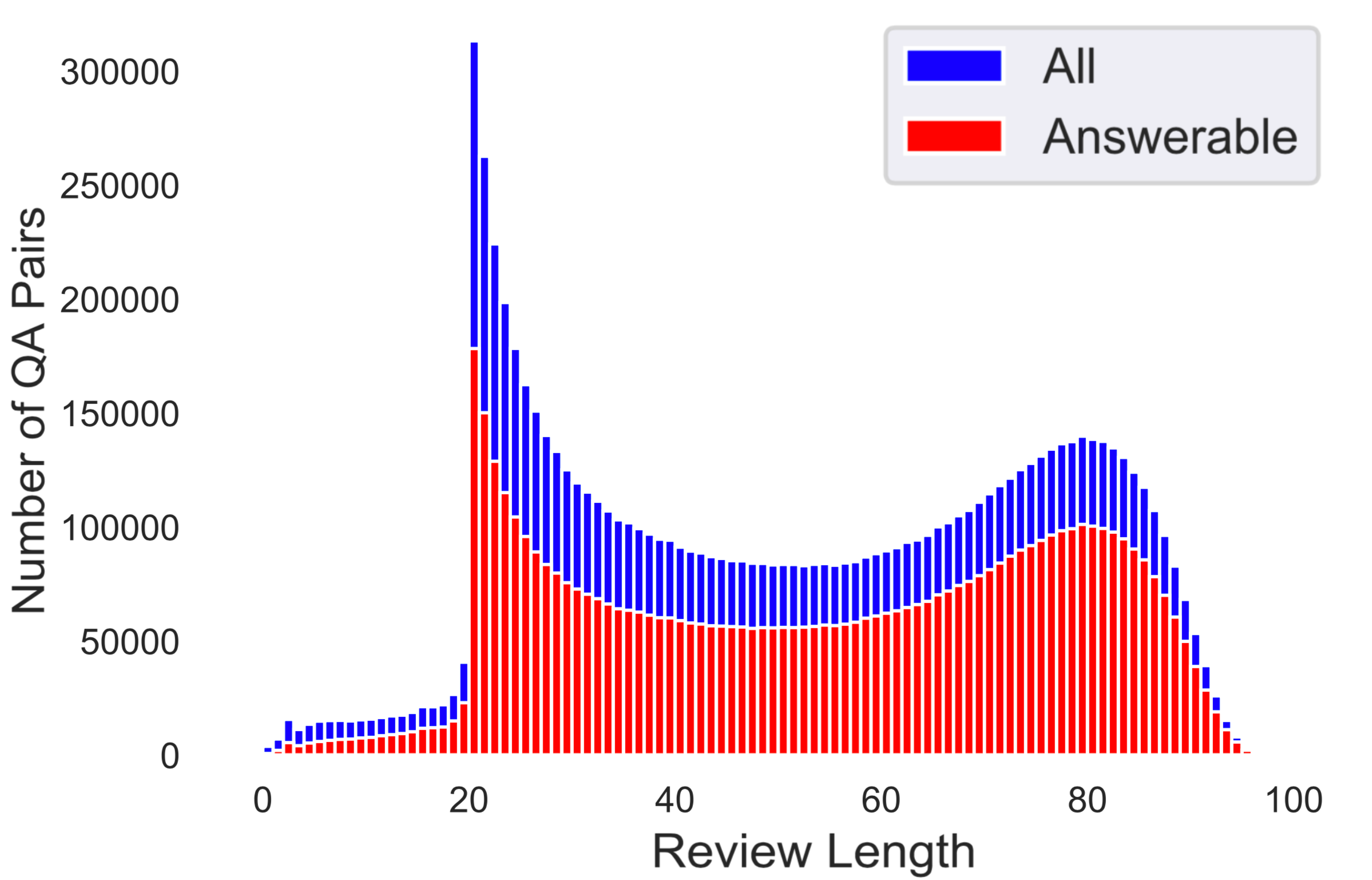} }
    \caption{Left to Right: Question length distribution, Answer length distribution and Review length distribution for Answerable Questions and All (Answerable + Non-Answerable) Questions.
    Note that the shape of distributions of \emph{All} questions and \emph{Answerable} questions is similar
    }
    \label{fig:lens}%
\end{figure*}

\subsection{Data Processing}\label{subsec:process}
\begin{figure}[ht]
\centering
    \includegraphics[scale=0.4]{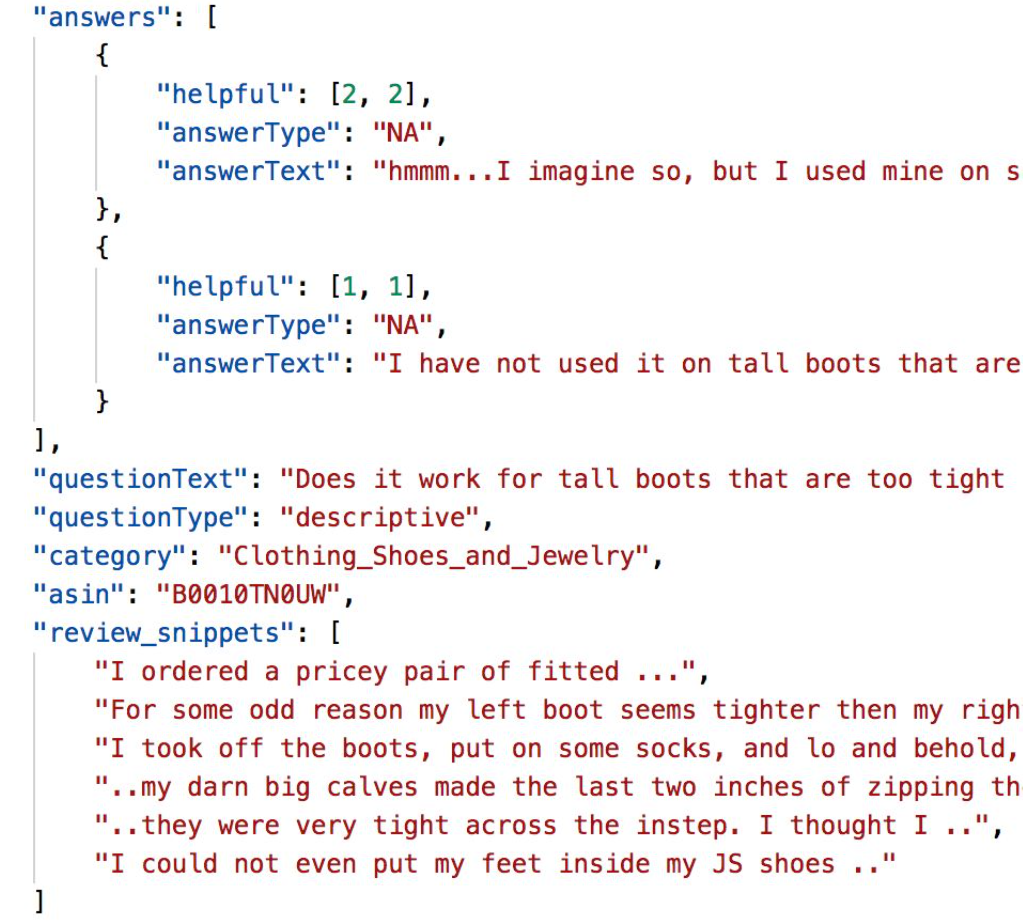}
    \vspace{-5px}
    \caption{A sample instance from the \emph{AmazonQA} dataset}
    \label{fig:example}
    \vspace{-5px}
\end{figure}

\begin{figure*}[ht]
    \centering
    \includegraphics[width=0.6\linewidth]{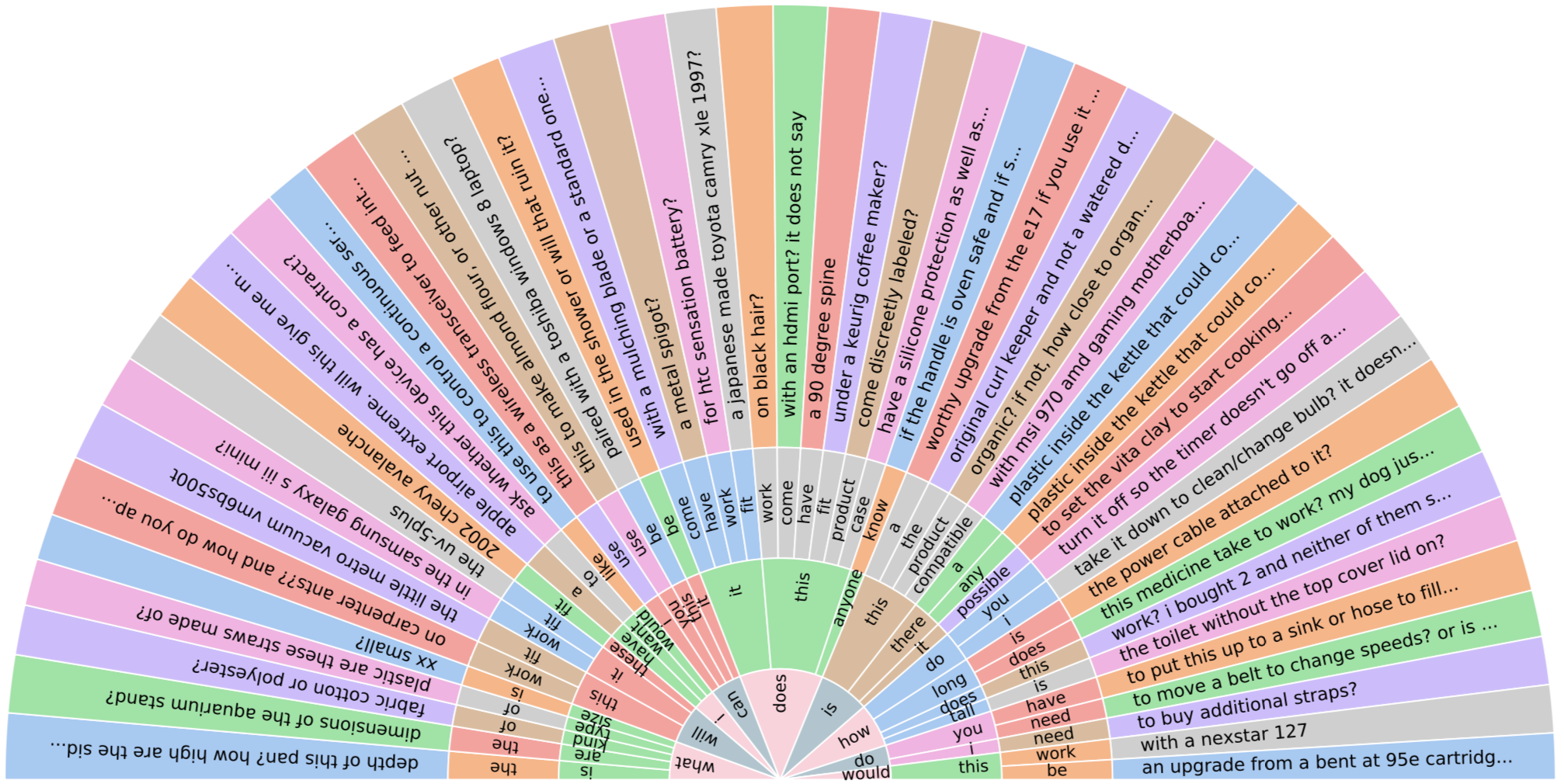}
    \caption{Distribution of most frequent first 3 words 
    in the questions of \emph{AmazonQA} dataset, with examples}
    \label{fig:donut}
\end{figure*}

As an artifact of web-crawling, many questions and reviews 
contain chunks of duplicate text that we identify and remove.
We find that a few of the reviews, questions, and answers,
are significantly longer as compared to their median lengths.
To mitigate this, we remove the outliers from the dataset. 
The distributions of questions, answers and reviews 
on the basis of length are shown in Figure \ref{fig:lens}.

Along with the raw product reviews, 
we also provide query-relevant review-snippets for each question.
We extract the snippets by first tokenizing the reviews, 
chunking the review text into snippets and 
ranking the snippets based on the TF-IDF metric. 
For tokenization, we remove all the capitalization 
except words that are fully capitalized as 
they might indicate abbreviations like `IBM'.
We consider the punctuation marks as individual tokens
(the only exception is apostrophe (') 
which is commonly used in words such as don't, I'll etc.,
that should not be separated). 
We then chunk the review text into snippets of length $100$, 
or to the end of a sentence boundary, whichever is greater.
These candidate snippets are then 
ranked on the basis of relevance between question 
and the review-snippet using BM25 score~\cite{bm25}. 
The set of 10 most relevant snippets 
for each question are provided in the dataset.



\subsection{Data Statistics} \label{subsec:data_stats}
We obtain roughly $923$k questions with $3.6$M answers 
on $156$k Amazon products having 14M unique reviews.
The average length of questions, answers and reviews 
is $14.8$, $31.2$ and $72.0$, respectively.

\begin{figure}[h]
\centering
    \includegraphics[width=0.9\linewidth]{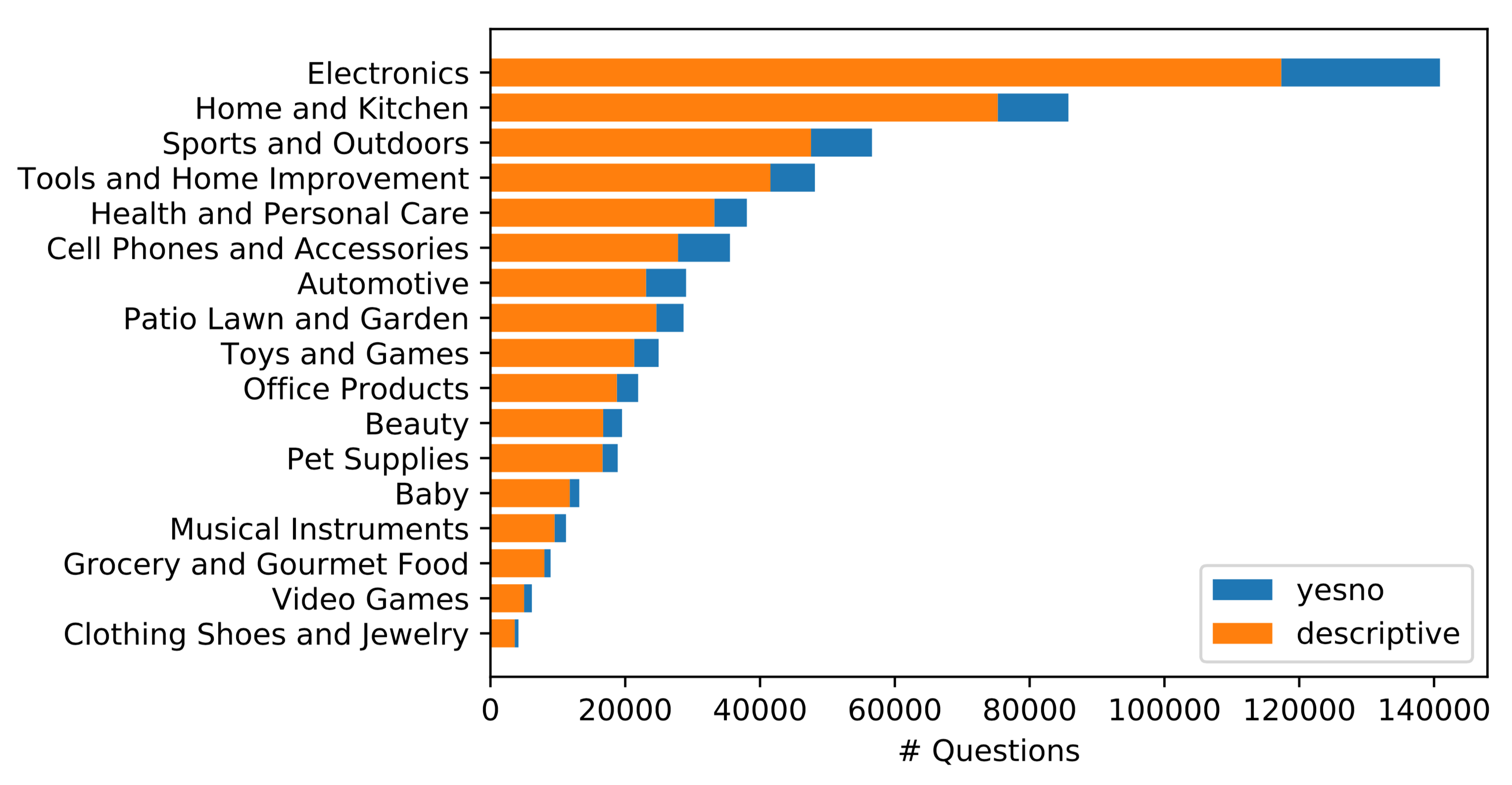}
    \vspace{-10px}
    \caption{Question Type Distribution}
    \label{fig:question_type}
    \vspace{-10px}
\end{figure}

\paragraph{Answerability Annotation}
We classify each question-context pair 
as \textit{answerable} or \textit{non-answerable} 
based on whether the answer to the question 
is at least partially contained in reviews. 
To train this \emph{Answerability classifier}, 
we obtain the training data through crowdsourcing using Amazon Mechanical Turk. 
The details of the Answerability classifier and the MTurk experiments 
are provided in \S\ref{sec:answer}. 
Our classifier marks roughly $570$K pairs 
as \emph{answerable} out of total of $923$K question with $72\%$ precision.
Note that the Fig \ref{fig:lens} shows similar shape of length distributions 
for both \emph{All} and \emph{Answerable} pairs, 
indicating that
the annotations cannot simply be predicted by the question/reviews lengths.
Further, there's no visible relation between the annotations and the answer lengths. 
\paragraph{Question Type Annotation}
\cite{mcauley2016addressing} classify the questions 
as \textit{descriptive} (open-ended) or \textit{yes/no} (binary). 
They use an regular expression based approach proposed by ~\cite{he2011summarization} 
Category-wise statistics of each class are shown in Figure \ref{fig:question_type}.
We also extract first three keywords of each question 
and provide an analysis of $50$ most frequent such $3$-grams in Figure \ref{fig:donut}. 
About $30\%$ of the questions are covered by these $3$-grams.




\subsection{Training, Development and Test Sets}
We split the data into train, development and test sets 
on the basis of products rather than on the basis of questions or answers. 
It means that all the QA pairs for a particular product 
would be present in exactly one of the three sets. 
This is to ensure that the model learns to formulate answers 
from just the provided reviews and not from other information about the product. 
After filtering the answerable questions, 
the dataset contains 570,132 question-answer-review instances. 
We make a random $80$-$10$-$10$ (train-development-test) split.
Each split uniformly consists of $85\%$ of \textit{descriptive} 
and $15\%$ \textit{yes/no} question types.
\section{Answerability}\label{sec:answer}
Since the QA pairs have been obtained independently from the context (reviews), 
there is no natural relationship between them. 
In order to ascertain that some part of the context actually answers the question, 
our system would require a classifier to determine 
whether a question-context pair is answerable or not. 
To do this, we conducted a crowdsourcing study 
on the \textit{Amazon Mechanical Turk} platform 
to obtain the labels on the question-context pairs 
which we use as the training data. 

\subsection{MTurk Experiments}
\begin{figure}[h]
		\centering
		\includegraphics[scale=0.36, angle=0, trim={3cm 0cm 0cm 0cm}, clip]{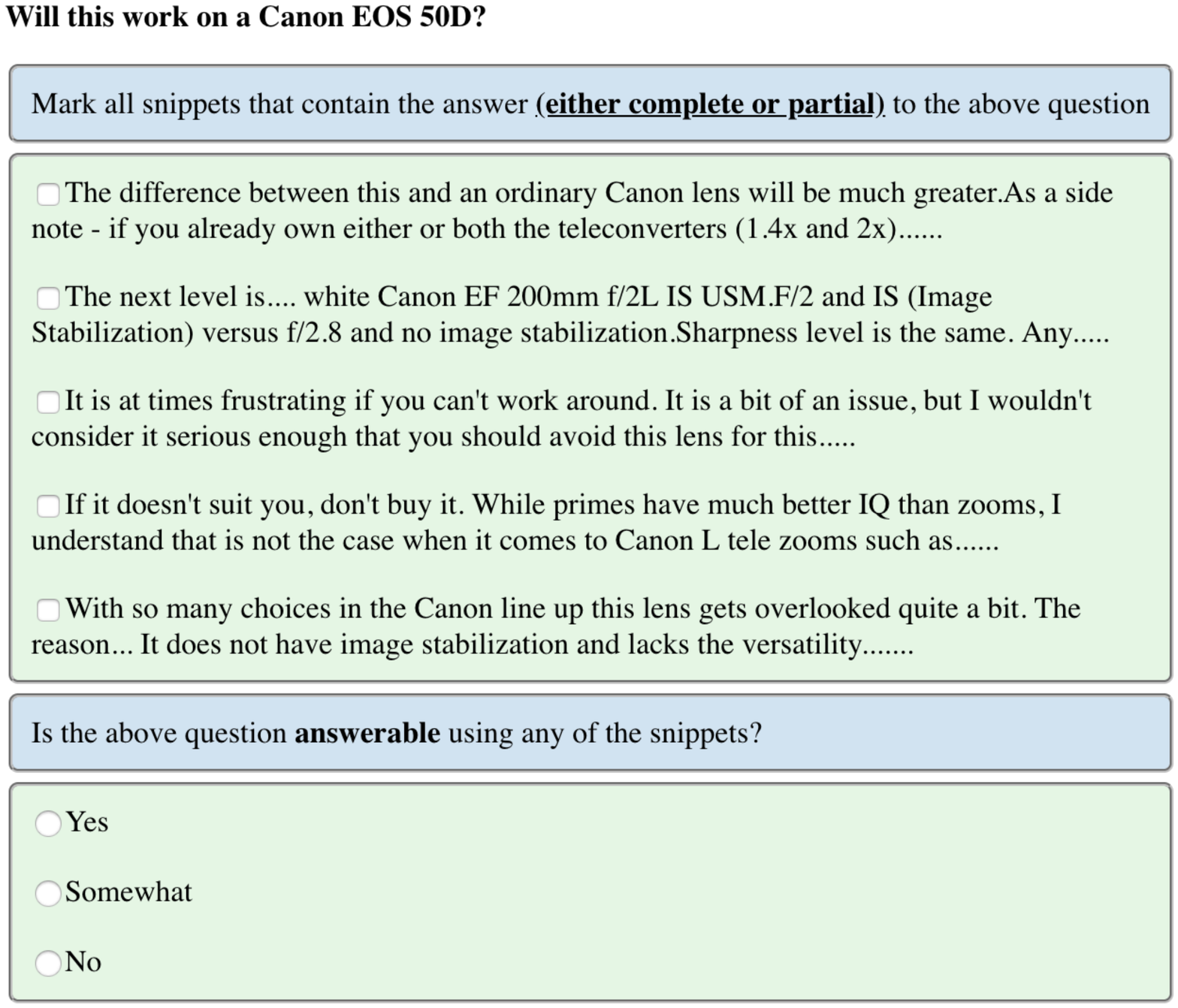}
		\vspace{-25px}
        \caption{Example of an AMT page with a question, corresponding review snippets, and the label to be marked}
        \label{fig:amt_page}
        \vspace{-5px}
\end{figure}

We provide workers with questions and a list of corresponding review snippets,
asking them to label whether the associated reviews 
are sufficient to answer each question. 
We experimented with multiple design variations 
on a sample set containing a couple hundred expert tagged examples 
before rolling out the full study and try to optimize for cost and accuracy. 
A single MTurk \emph{hit} is a web page shown to the workers,
consisting of $N=5$ questions,
out of which one of them is a \textit{`decoy'} question.
\textit{Decoy} questions are a set of expert annotated, 
relatively easier samples which have $100\%$ inter-annotator agreement. 
The responses to the decoy questions allow us 
to compute a lower bound of a worker's performance.
With this guiding metric, we detail our most efficient design template below.
An example question with corresponding evidence in review text is provided in Table \ref{tab:amt_span}.
\begin{itemize}[noitemsep,topsep=3pt]
    \item In addition to the \textit{answerable} (Y) and \textit{not-answerable} (N) labels, 
    we also provided a \textit{somewhat} (S) label to the workers. 
    \textit{Somewhat} indicates that the reviews contain information 
    that partially answers the question. 
    This is helpful to provide a provisional answer with limited insight, 
    rather than a complete answer.
    \item Workers are required to mark snippets that (partially/completely) answer the question. Although we do not at present leverage this information,
    we observe that this inclusion leads workers to peruse the reviews, 
    consequently minimizing mislabeling.
    \item Each hit has a total of $5$ questions, including a decoy question. 
    To discard careless workers, 
    we estimate \textit{Worker Error} as follows $\sum abs(\textit{worker\_label} - \textit{true\_label}) / 2.0$. 
    The label values are mapped as $\{Y : 1, S: 0, N : -1\}$
\end{itemize}
Examples of labels with the corresponding evidence 
are shown in Figure \ref{fig:amt_page}.

\begin{table}[h]
		\centering
		\small
        \begin{tabular}{cc}
        \toprule
        \multicolumn{2}{c}{Q: What is the dimension of the product?}\\
        \midrule
        \textbf{Span} & \textbf{Is Ans?} \\
        \midrule
        ``...the size of the bag is &\\about 11in x 12in..."  & Yes\\
        \hline
        ``...this bag is big enough &\\to fit a macbook..."  & Somewhat\\
        \hline
        ``...the size of the front & \\pocket is very small..."  & No\\ 
        \bottomrule
        \end{tabular}
        \caption{An example question (\textit{Q}) with evidence in review text (\textit{Span}) and expected label (\textit{Is Ans?})}
        \label{tab:amt_span}
\end{table}


Finally, we rolled out the study on a total of $6000$ (non-decoy) questions, 
and eliminated the workers that performed poorly on the decoy questions. 
We retained 3,297 labeled questions after filtering. 
On an average it took $\sim8.6$ minutes for workers to complete one hit.

\subsection{Answerability Classifier}
We process the data and extract query-relevant review-snippets 
as per \S\ref{subsec:process}. 
We use a sentence tokenizer on the top $5$ review snippets 
to generate a list of sentences used as context to ascertain 
whether the question is answerable. 
To featurize the question-context pairs,
we obtain two kinds of representations for the sentences, 
a corpus based tf-idf vectorizer and a tf-idf weighted Glove embeddings \cite{pennington2014glove} to capture both statistical and semantic similarity 
between question and snippets.

As features, we use the number of question tokens, 
number of context tokens, absolute number 
and the fraction of question tokens 
that are also present in the context. 
We also use the cosine similarity between mean and max pooled representations
of sentences with that of the question as features, 
thus getting a total of $8$ features for each item.

We split the dataset into train, validation and test sets 
with $2967$, $330$ and $137$ question-context instances, respectively. 
We train a binary classifier on the data collected from MTurk, 
where the instances with labels as \textit{Yes} and \textit{Somewhat} 
are labeled as positive instances, 
and \textit{No} are labeled as negative instances.
The test set is annotated by 4 experts 
and we use it to measure the performance of MTurk workers, 
the binary classifier as well as the inter-expert agreement.
The results are shown in the Table \ref{tab:ansResults}.
We use this model to classify answerability for the whole dataset.
\begin{table}[t!]
    \small{}
    \centering
    \begin{tabular}{c c c c c} 
        \toprule
        & Precision & Recall & F1-Score \\
        \midrule
        Expert  &  $0.92$ & $0.81$ & $0.86$ \\
        Worker & $0.73$ & $0.73$ & $0.73$ \\
        Classifier & $0.67$ & $0.83$ & $0.74$\\
        \bottomrule
    \end{tabular}
    \caption{Answerability precision, recall and F1-score by the expert, worker and the Logistic Regression classifier (C=1, threshold=0.6)}
    \label{tab:ansResults}
    \vspace{-5px}
\end{table}
\section{Baseline Models} 
\label{sec:models}
To benchmark the generation of natural language answers, 
we implement three language models. 
We also implement reading-comprehension (RC) models 
that perform well on exiting span-based QA datasets 
to assess their performance on this task.

\subsection{Language Models}
To evaluate our ability to generate answers given reviews, 
we train a set of models for both answer generation (language modeling)
and conditional language modeling (sequence to sequence transduction). 
If $a$ is an answer, $q$ is the corresponding question,
and $R$ is a set of reviews for the product, 
we train models to approximate the conditional distributions: 
$P(a)$,  $P(a \mid q)$ and $P(a \mid q, R)$. 

We train a simple language model 
that estimates the probability of an answer, $P(a)$, via the chain rule. 
The other two models estimate the probability of an answer 
conditioned on (i) just the question and (ii) both the question and the reviews. 
By evaluating models for all three tasks,
we can ensure that the models are truly making use 
of each piece of additional information.
Such ablation tests are necessary, 
especially in light of recent studies 
showing several NLP tasks to be easier than advertised
owing to some components being unnecessary 
\cite{kaushik2018much,gururangan2018annotation}.

\begin{figure}[h]
\centering
    \includegraphics[scale=0.5]{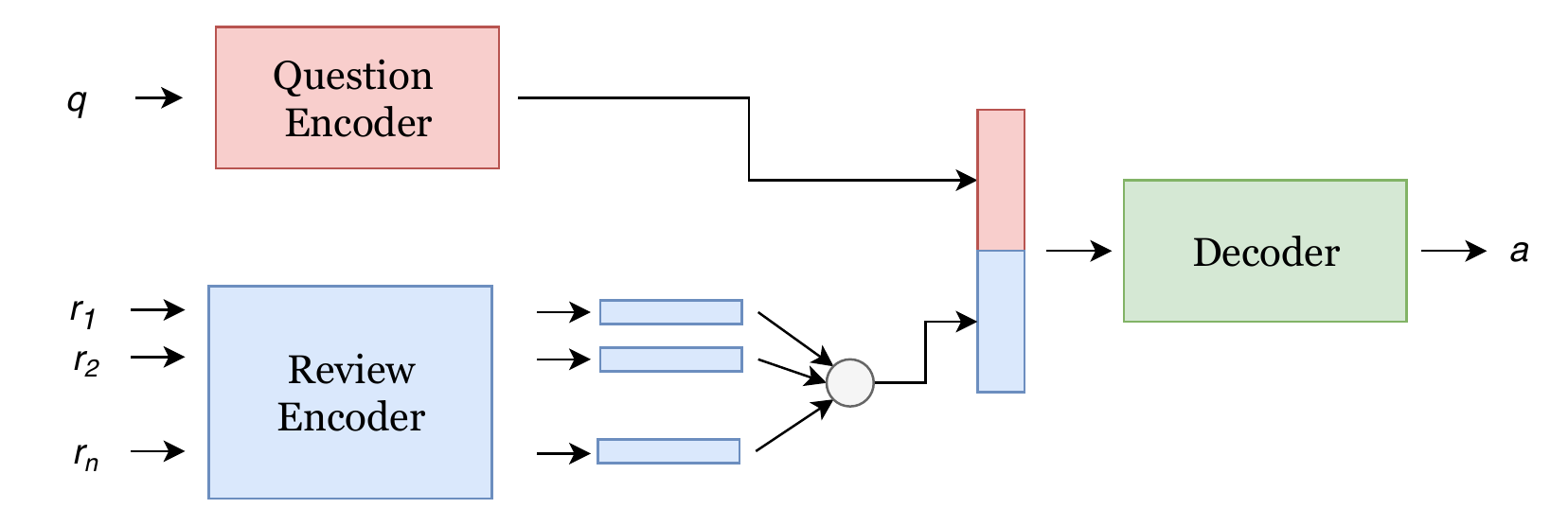}
    \caption{A schematic overview of the model $P(a \mid q, R)$. The $P(a \mid q)$ and $P(a)$ models are special cases where the review representation or both review and question are absent. All encoders and decoders are LSTM-based.}
    \label{fig:lmseq2seq}
    \vspace{-10pt}
\end{figure}

The three language models not only provide us an insight 
into the difficulty of predicting an answer 
using the question and reviews but also act as natural baselines 
for generative models trained on this dataset.
To implement these language models, 
we use a generalized encoder-decoder based 
sequence-to-sequence architecture (Figure \ref{fig:lmseq2seq}).
The reviews are encoded using an LSTM-based encoder, 
and the encoded representations are averaged 
to form an aggregate review representation. 
The question representation (also by an LSTM-based encoder) 
and the aggregated review representation are concatenated 
and used to initialize an LSTM-decoder
that generates the tokens of the answer at each step. 
\subsection{Span-based QA Model} \label{subsec:spanbased}
To assess the performance of span-based RC models on our dataset, 
we convert our dataset into a span-based format.
To do that, we use the descriptive answers from users 
to heuristically create a span (sequences of words) 
from the reviews that best answer the question. 
We then train a span based model, R-Net \cite{mcr},
that uses a gated self-attention mechanism and pointer networks
to find the location of the answers in the reviews.
\paragraph{Span Heuristics} 
To create spans for supervision, 
we first create a set of candidate spans
by either considering (i) all $n$-grams,
where $n \in \{10, 20\}$, 
or (ii) all sentences in the reviews. 
We then rank the spans based on heuristics such as 
(i) BLEU-$2$ \cite{papineni2002bleu} or ROUGE \cite{lin2004rouge} 
with the actual answers, and 
(ii) BM25 based IR score match with the question. 
To evaluate span quality, 
we annotate a test sample of $100$ questions 
and corresponding spans from each heuristic as \textit{answerable}
and \textit{not-answerable} using the spans. 
From this evaluation (Table \ref{tab:span_heuristic_evaluation}), 
we identify IR- and BLEU-$4$-based sentences 
as the most effective span heuristics. 
Even though the fraction of questions \textit{answerable} 
using the spans is relatively small, 
the spans serve as noisy supervision for the models 
owing to the large size of the training data. 
In table \ref{tab:span_based}, we show performance 
of R-Net when trained on spans generated by two of these span heuristics.

\begin{table}[h]
    \centering
    \small{}
    \begin{tabular}{c c c} 
        \toprule
            Span Heuristic & \multicolumn{2}{c}{fraction of \textit{answerable} questions} \\
            & sentences & $n$-grams \\
            \midrule
            IR      & $0.58$  & $0.24$ \\
            BLEU-$2$  & $0.44$  & $0.21$ \\
            BLEU-$4$  & $0.45$  & $0.26$ \\
        \bottomrule
    \end{tabular}
    \caption{Expert evaluation of span heuristics on test set. 
    }
    \label{tab:span_heuristic_evaluation}
    \vspace{-5px}
\end{table}
\section{Results and Evaluation} \label{sec:results}
\begin{table*}[t]
    \renewcommand\thetable{7}
    \centering
    \small{}
    \begin{tabular}{ l c c c c c c c c c} 
        & \textbf{Bleu-1}   & \textbf{Bleu-2}   & \textbf{Bleu-3}   & \textbf{Bleu-4}    & \textbf{Rouge} & $P(a \mid q)$ / $P(a \mid q, R)$ \\ 
        \toprule
        \textbf{Heuristic Baselines}  & & & & & & \\
        \hspace{4mm} A sentence from reviews as answer & & & & & & \\ 
        \hspace{7mm} Random sentence& $78.56$ & $63.95$ & $44.37$ & $29.87$ & $49.12$ &  $97.27$ / $88.51$ \\ 
        \hspace{7mm} Top-1 using IR & $89.49$ & $74.80$ & $56.76$ & $43.52$ & $61.48$ & $128.32$ / $115.66$ \\ 
        \hspace{7mm} Top-1 Using BLEU & $\mathbf{92.74}$ & $\mathbf{78.43}$ & $\mathbf{60.91}$ & $\mathbf{48.08}$ & $\mathbf{62.68}$ & -\\
        \hspace{4mm} Entire review as answer & & & & & & \\ 
        \hspace{7mm} Top-1 (helpfulness score)  & $20.66$ & $19.78$ & $16.39$ & $12.54$ & $37.01$ & $146.00$ / $267.86$\\ 
        \hspace{7mm} Top-1 (Wilson score)       & $20.74$ & $19.84$ & $16.44$ & $12.58$ & $37.26$ & $144.44$ / $126.07$ \\ 
        \hline 
        \textbf{Neural Baseline}  & & & & & & \\
        \hspace{4mm} R-Net  & & & & & & \\ 
        \hspace{7mm} BLEU Heuristic   & $47.04$ & $40.32$ & $31.48$ & $23.92$ & $40.22$ & - \\ 
        \hline
        \textbf{Human Answers}  & & & & & & \\ 
        \hspace{4mm} Amazon User Community     &   $80.88$ & $68.86$ & $54.36$ & $42.01$ & $62.18$ & -\\ 
        \hspace{4mm} Expert (Spans)          & $68.33$ & $57.79$ & $44.61$ & $34.43$ & $51.09$ & -\\ 
        \hspace{4mm} Expert (Descriptive)    & $53.67$ & $46.56$ & $37.81$ & $30.76$ & $53.31$ & -\\ 
        \bottomrule
    \end{tabular}
    \caption{The performances and \emph{perplexities} of various methods on the \emph{AmazonQA} test set. The neural baseline, R-Net, is trained using spans created from BLEU Heuristic as explained in \ref{subsec:spanbased}}
    \label{tab:baseline_results}
    \vspace{-10px}
\end{table*}

\paragraph{Language Models}
For the language models, we compare the validation perplexities 
for all the three models (\S\ref{sec:models}) 
on the test dataset of \emph{AmazonQA}. 
Since perplexity is the exponential 
of the negative log likelihood per each word in the corpus, 
a lower perplexity indicates that the model deems the answers as more likely. 
Naturally, we expect the perplexities of the model 
$P(a \mid q, R)$ to be lower than those of $P(a \mid q)$, 
which should, in turn, be lower than perplexities from $P(a)$. 
We empirically verify this on the test set of AmazonQA (table \ref{tab:results-ppl}). 

\begin{table}[H]
    \renewcommand\thetable{6}
    \small{}
    \centering
    \begin{tabular}{ c c }
        \toprule
        {Model} & Test Perplexity \\
        \midrule
        $P(a)$ & $97.01$ \\
        $P(a \mid q)$ &  $70.13$ \\
        $P(a \mid q, R)$ & $65.40$ \\
        \bottomrule
    \end{tabular}
    \caption{Perplexities of the language models on the test set
    }
    \label{tab:results-ppl}
    \vspace{-5px}
\end{table}

\paragraph{Heuristic-based answers}
For comparison with our baseline models, 
we consider the following trivial heuristic-based 
experiments to predict the answer:
(i) \textbf{A sentence from the reviews as answer}: 
We consider the top-ranked sentence 
based on IR and BLEU-$2$ scores 
as an answer to the question.
We also experiment with using a random sentence as an answer;
(ii) \textbf{An entire review as the answer}: 
We use the top review based on 
(i) \textit{helpfulness} of the reviews, as indicated by Amazon users; and (ii) {\textit{Wilson Score}}\footnote{Lower bound of Wilson score 95\% CI for a Bernoulli parameter} \cite{agresti1998approximate} 
derived from \textit{helpfulness} and 
\textit{unhelpfulness} of the reviews, as the answer. 

While these heuristics are largely similar 
to the span heuristics described in \ref{subsec:spanbased},
the difference is that here, 
each of these heuristics is treated as a "QA model" in isolation
for comparison with more complex models
while the span heuristics are used
to generate noisy supervision
for training a span-based neural model.

\paragraph{Metrics}
As the answers to both \textit{yes/no} and \textit{descriptive} questions 
are descriptive in nature, we use BLEU and ROUGE scores as the evaluation metrics.
We use the answers provided by the Amazon users as the reference answer. 

\paragraph{Human Evaluation} 
We compare to human performance for both experts and Amazon users. 
For expert evaluation, we annotate a test set of 100 questions 
for both span-based and descriptive answers. 
To compute the performance of Amazon users, for each question, 
we evaluate an answer using the other answers 
to the same question as reference answers.

\begin{table}[H]
    \renewcommand\thetable{5}
    \centering
    \small{}
    \begin{tabular}{c c c} 
        \toprule
        Span Heuristic & Exact Match & F1 \\ 
        \midrule
        BLEU\_$2$ & $5.71$ & $33.97$ \\
        ROUGE & $5.94$ & $30.35$ \\
        \bottomrule
    \end{tabular}
    \caption{The test-set performance of R-Net using different span generating heuristics for supervision}
    \label{tab:span_based}
    \vspace{-5px}
\end{table}

Table \ref{tab:baseline_results} shows the performances 
of the different baseline models on our dataset. 
We note that the scores of the answers from Amazon users 
as well as the sentence-based heuristic baselines 
are higher than the those of the span-based model (R-net). 
This indicates a large scope of improvement for the QA models on this dataset.

Note that a random sentence from the review 
performs nearly as well as answers provided by Amazon users. 
This might be due to the fact that the Amazon users 
can see the existing answers and would be more inclined 
to write a new answer to add additional information. 
This also partially explains the high scores for IR-based top-sentence heuristic. 
Since many of the questions are of the type \textit{yes/no}, 
the conventional descriptive answer evaluation metrics such as BLEU and ROUGE, 
that are based on token similarity may not be the best metrics for evaluation. 
A better way to evaluate the system would be to have 
a combined set of metrics such as accuracy for \textit{yes/no} questions, 
BLEU and ROUGE for descriptive and perplexities from the language models. 

%
%
%
\section{Conclusion}
We present a large Question Answering dataset, 
that is interesting for the following reasons:
i) 
the answers are provided by users in a real-world scenario;
ii) the questions are frequently only partially answerable based on the reviews, leaving the challenge to provide the best answer under partial information; and
iii) the dataset is large, comprising several categories and domains thus possibly useful for learning to answer out-of-domain questions.
To promote research in this direction, 
we publicly release\footnote{{https://github.com/amazonqa/amazonqa}} the dataset and
our implementations of all baselines. 

\section*{Acknowledgments}
We thank Adobe Experience Cloud for their generous support of this research through their Data Science Research Award.

\balance


\bibliographystyle{named}
\bibliography{references}

\end{document}